\documentclass[letter]{spie}  

\usepackage{amsmath,amsfonts,amssymb}
\usepackage{graphicx}
\usepackage[colorlinks=true, allcolors=blue]{hyperref}
\usepackage{color}
\usepackage{graphicx}
\usepackage{subcaption}
\usepackage[english]{babel}

\title{Direct Prediction of Cardiovascular Mortality from Low-dose Chest CT using Deep Learning}

\author[a]{Sanne G.M. van Velzen}
\author[a]{Majd Zreik}
\author[a]{Nikolas Lessmann}
\author[a]{Max A. Viergever}
\author[b]{Pim A. de Jong}
\author[c]{Helena M. Verkooijen} 
\author[a]{Ivana I\v sgum}
\affil[a]{Image Sciences Institute, University Medical Center Utrecht, the Netherlands}
\affil[b]{Department of Radiology, University Medical Center Utrecht, the Netherlands}
\affil[c]{Imaging Division, University Medical Center Utrecht, Utrecht University, the Netherlands}

\begin{document} 
\maketitle

\begin{abstract}
Cardiovascular disease (CVD) is a leading cause of death in the lung cancer screening population. Chest CT scans made in lung cancer screening are suitable for identification of participants at risk of CVD. Existing methods analyzing CT images from lung cancer screening for prediction of CVD events or mortality use engineered features extracted from the images combined with patient information. In this work we propose a method that automatically predicts 5-year cardiovascular mortality directly from chest CT scans without the need for hand-crafting image features.

A set of 1,583 participants of the National Lung Screening Trial was included (1,188 survivors, 395 non-survivors). Low-dose chest CT images acquired at baseline were analyzed and the follow-up time was 5 years. To limit the analysis to the heart region, the heart was first localized by our previously developed algorithm for organ localization exploiting convolutional neural networks. Thereafter, a convolutional autoencoder was used to encode the identified heart region. Finally, based on the extracted encodings subjects were classified into survivors or non-survivors using a support vector machine classifier. The performance of the method was assessed in eight cross-validation experiments with 1,433 images used for training, 50 for validation and 100 for testing. The method achieved a performance with an area under the ROC curve of 0.72. 

The results demonstrate that prediction of cardiovascular mortality directly from low-dose screening chest CT scans, without hand-crafted features, is feasible, allowing identification of subjects at risk of fatal CVD events.
 
\end{abstract}

\section{PURPOSE}
\label{sec:intro}  

Besides lung cancer, cardiovascular disease is a leading cause of death in the lung cancer screening population\cite{national2011reduced}\spacefactor\sfcode`\.{}. Moreover, it has been shown that chest CT scans used for lung cancer screening are suitable for identification of participants at risk of cardiovascular disease (CVD)\cite{jacobs2012coronary,chiles2015association,mets2013lung,de2015automatic}\spacefactor\sfcode`\.{}. Previous methods that investigated prediction of CVD events and all-cause mortality used known quantitative CVD image markers and combined them with subject data. Using weighted Cox proportional hazards regression, Chiles et al.\cite{chiles2015association} showed that quantitative as well as visually assed coronary artery calcium (CAC) scores extracted from screening low-dose chest CT are predictive for CVD and all-cause mortality in the National Lung Screening Trial (NLST). Similarly, Mets et al.\cite{mets2013lung} used Cox regression based on semi-automatically detected CAC scores and thoracic aorta calcium (TAC) volume as well as subject data, to perform prediction of CVD events and all-cause mortality in the Dutch-Belgian lung cancer screening trial (NELSON). De Vos et al. \cite{de2015automatic} performed prediction of CVD events in the same population using a support vector machine (SVM) classifier that employed automatically extracted CAC and TAC scores as features. 


These approaches relied on hand-crafted features that are already established as CVD biomarkers. However, besides these known biomarkers, chest CT scans may contain yet unknown features predictive of CVD mortality. Hence, we propose a method based on unsupervised feature learning which is able to automatically predict CVD mortality directly from chest CT scans and is therefore not limited to known quantitative image markers related to CVD. 


\section{DATA}
\label{sec:data}
This study included 1,583 participants of the National Lung Screening Trial. NLST included current and former heavy smokers between the age of 50 and 74\cite{national2011reduced}\spacefactor\sfcode`\.{}. All 395 participants who died of CVD within 5 years from acquisition of the baseline CT scan (non-survivors) were included. In addition, 1,188 participants who were still alive after this period (survivors) were randomly selected.

For each subject a CT scan acquired at baseline was analyzed. Low-dose chest CT scans were made with breath-hold, without contrast enhancement and without ECG synchronization.  Scans were acquired using helical scanning mode and a tube voltage of 120 kVp or 140 kVp, depending on the subject's weight. In-plane resolution ranged from 0.49 mm to 0.98 mm with a slice thickness between 1.0 mm and 2.5 mm. The scans were acquired at 32 different medical centers with 13 different scanner models.

\section{METHODS}
\label{sec:methods}
To investigate whether analysis of the heart visualized in chest CT enables prediction of CVD mortality, the method first extracts a bounding box around the heart. This is done with our previously designed and trained algorithm\cite{de2017convnet} that employs a CNN to determine the presence of the heart in axial, coronal and sagittal slices of the chest CT image and subsequently combines these to define a 3D bounding box around the heart. Thereafter, to ensure equal image resolution in our data set, we resample all extracted heart volumes to isotropic resolution of 1.0 mm. Moreover, to enhance differentiation among soft tissues in the heart (e.g. fat, muscle) and to preserve influence of high density structures (e.g. CAC, TAC), extracted volumes are clipped between [-160, 840] HU. 

Because of the relatively low number of the available samples, prediction using e.g. a CNN that would extract features and perform classification of subjects into survivors and non-survivors was not feasible. Therefore, similar to the work of Zreik et al.\cite{zreik2018deep}\spacefactor\sfcode`\.{}, a convolutional autoencoder (CAE) is used to encode the volumes containing the heart in an unsupervised fashion. Thereafter, a conventional machine learning classifier exploiting the extracted encondings is employed to classify subjects into survivors and non-survivors. 

\begin{figure}[t]
   \begin{center}
   \begin{tabular}{c} 
   \includegraphics[height=3.3cm]{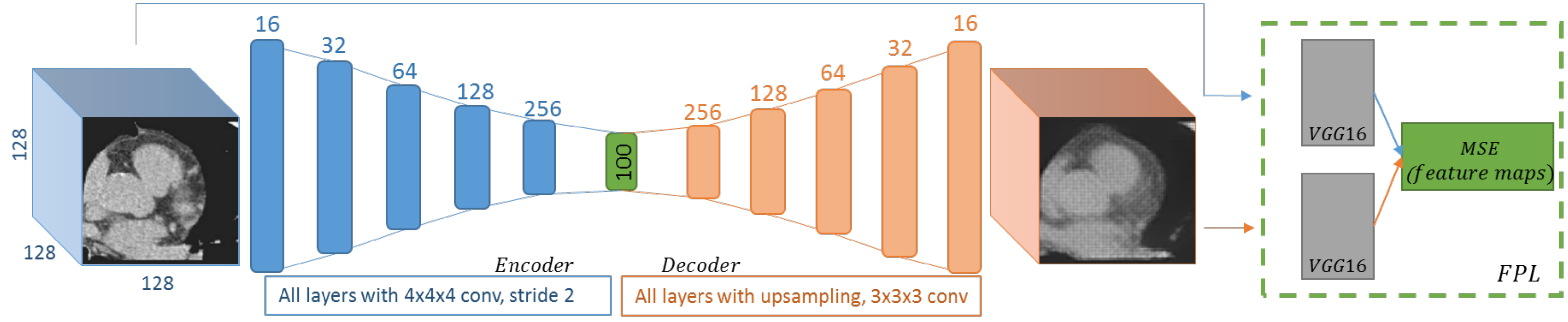}
   \end{tabular}
   \end{center}
   \caption[example] 
   {\small{ \label{fig:arch} 
CAE with feature perceptual loss (FPL): Input images are encoded into 100 encodings and reconstructed by the decoder. Both the input and reconstructed images are fed into a fixed VGG16 network, where the FPL is determined by the MSE over the feature maps derived from the input image and the reconstructed image.}}
   \end{figure}

The CAE used in this work (Figure \ref{fig:arch}) consists of an encoder that compresses the images into representative encodings and a decoder that reconstructs the images during training. The encoder analyzes images, cropped to the heart volume, zero-padded to $128\times128\times128$ voxels and consists of 5 convolutional layers with $4\times4\times4$ kernels. A stride of 2 was applied to achieve spatial downsampling without the need for deterministic spatial functions such as max-pooling. The encoder ends in a dense layer with 100 units, which represents the encodings vector. The decoder consists of 5 upsampling layers and 5 convolutional layers with $3\times3\times3$ kernels and a stride of 1. All convolutional layers are followed by batch normalization and LeakyReLu activation ($\alpha = 0.3$).

Typically, in training the CAE, the mean squared error (MSE) between the reconstructed image and the original image is used as a loss function. However, to capture the contrast among soft tissues in low-dose CT without intravenous contrast enhancement, in this work the loss function of the CAE is defined employing the feature perceptual loss (FPL), which captures perceptual differences and spatial correlations better\cite{hou2017deep}\spacefactor\sfcode`\.{}. To compute the FPL, both the input image and reconstructed image are separately fed into a fixed VGG16\cite{simonyan2014very} network pretrained on ImageNet (Figure \ref{fig:arch}). The FPL is then defined as the MSE between the feature maps in this network derived from the input image and the reconstructed image. Because VGG16 is a network designed for 2D images, the loss was calculated over 2D axial slices of the 3D image volumes.

Thereafter, three different classifiers are trained using the encodings obtained with the CAE: a neural network (NN), a random forest classifier (RFC) and a support vector machine (SVM) classifier. For the RFC and SVM a grid search on the validation set is performed to find the optimal parameter settings.


\section{EXPERIMENTS AND RESULTS}

To assess the performance, eight cross-validation experiments were performed. In each experiment, 100 images were selected as test set, 50 images were selected as validation set and the remaining 1,433 images were used as training set. Test and validation sets were balanced with respect to classes. In the test sets, images of the non-survivors were sampled such that each non-survivor was included once in the test sets. Images of the survivors were randomly selected from the available set.



To augment the training set for the CAE, random rotations around all three image axes were used. The angle of rotation was randomly chosen from a normal distribution with a mean of 0 and a standard deviation of 10 degrees. The CAE was trained in 100,000 iterations, with a batch size of 2 images zero-padded to the required input size. The Adam optimization algorithm was used with a learning rate of 0.001.

To classify subjects into survivors and non-survivors, the classifiers were trained using the encodings obtained with the CAE. The NN (6 units dense layer and 2 units output, dropout p=0.5, categorical crossentropy loss, lr=0.0001) was trained in 25,000 iterations with balanced batches of 100 examples, using the Adam optimization algorithm. The RFC consisted of 75 trees and for the SVM the $\gamma$ and $c$ were set to 0.0001 and 100, respectively.

To determine the influence of FPL on the reconstructions, we trained an additional CAE with standard MSE as loss function. Using MSE as a loss function resulted in a mean absolute error of 19 ($\pm$ 5) HU and training with FPL resulted in mean absolute error of 20 ($\pm$ 6). While the mean absolute reconstruction errors are similar, the CAE trained with FPL learned sharper contrast between structures, as shown in Figure \ref{fig:num_enc}A. 
 

In each cross-validation experiment a CAE was trained and the classifiers were evaluated. The performance of the method was assessed with a receiver operating characteristic (ROC) curve. The SVM achieved the best performance with an area under the curve (AUC) of 0.72 ($\pm$ 0.07 standard deviation, Figure \ref{fig:num_enc}B). The ROC of the neural network and the RFC had an AUC of 0.71 ($\pm$ 0.06) and 0.70 ($\pm$ 0.06), respectively.

\section{DISCUSSION AND CONCLUSION}
In this work, a method for prediction of cardiovascular mortality from lung screening chest CT scans has been proposed. Unlike previous predictive models, the proposed method does not use hand-crafted image features, but performs prediction directly from the images containing the heart. 

\begin{figure}[t]
   \begin{center}
   \begin{tabular}{c} 
   \includegraphics[height=4.38cm, trim={0 6mm 0 2mm},clip]{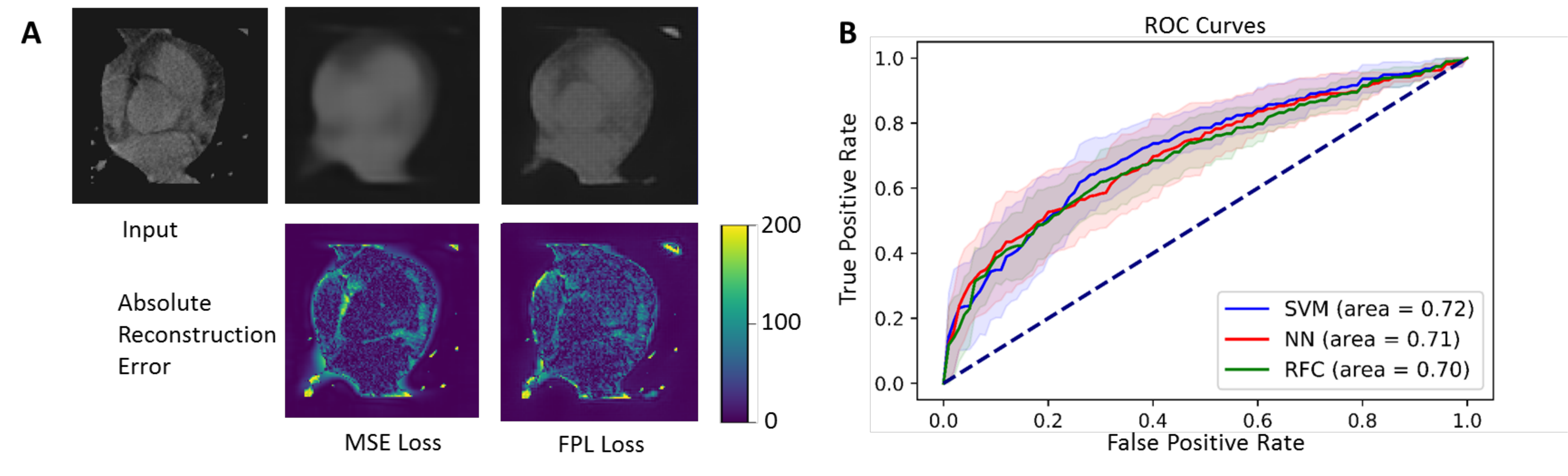}
   \end{tabular}
   \end{center}
   \caption[example] 
   {\small{ \label{fig:num_enc} 
A) Input and reconstructed images generated by CAEs trained with MSE loss and FPL (top), with corresponding reconstruction error maps (bottom). B) ROC curves for prediction with SVM, NN and RFC, with standard deviation over 8 cross-validation experiments.}}
   \end{figure}

The experiments show that the CAE using FPL preserves structures likely containing important information for prediction of CVD mortality, such as the coronary arteries, aorta, and fat around the heart (Figure \ref{fig:num_enc}A). The CAE was trained to reconstruct heart images and was agnostic to the subsequent classification task. It would be interesting to investigate end-to-end training where the CAE is trained while optimizing for a subsequent classification task. Besides likely improvement in the performance by end-to-end training, such an approach might allow identification of image areas important for prediction and thereby allow the confirmation of known and possibly the discovery of novel image markers of CVD mortality. Alternatively, a CNN could be employed to perform classification directly. However, our preliminary experiments showed that end-to-end training and direct classification require a larger data set, which is often not available.

The three evaluated classifiers show a similar performance for prediction of CVD mortality based on the extracted encodings. Furthermore, the presented method shows a similar performance to previous methods proposed by Mets et al. and De Vos et al. (AUC = 0.71)\cite{mets2013lung,de2015automatic}, that describe prediction of CVD events. This is remarkable since the CAE is unsupervised and does not incorporate prior knowledge about the image, the subsequent classification task or relevant biomarkers. Moreover, the proposed method achieved similar performance to work by Mets et al. and De Vos et al.\cite{mets2013lung,de2015automatic}\spacefactor\sfcode`\.{}, without adding subject data like age, smoking history or sex, as they proposed. However, comparison with these methods is somewhat limited by different outcome definitions and populations. Although image analysis without additional subject data may be simpler in application, future work could investigate whether incorporating subject data in the proposed method would improve the performance.

In conclusion, this work demonstrates that the prediction of cardiovascular mortality directly  from low-dose screening chest CT scans is feasible. This might allow identification of subjects undergoing lung screening with CT who are at risk of fatal CVD events and who might benefit from preventive treatment.

\section{NEW OR BREAKTROUGH WORK}
A machine learning system is presented that is able to predict cardiovascular death within five years from a low-dose chest CT scan, without prior information about the image or subjects. 

\small{\acknowledgments{The authors thank the National Cancer Institute for access to NCI's data collected by the National Lung Screening Trial. The statements contained herein are solely those of the authors and do not represent or imply concurrence or endorsement by NCI. The authors hereby state that this work is not submitted elsewhere.}}

{\footnotesize
\bibliography{prediction}} 

\bibliographystyle{spiebib} 

\end{document}